\documentclass[twocolumn,cleanfoot,10pt]{asme2ej}
\usepackage{lineno,hyperref}
\modulolinenumbers[5]

\usepackage{amssymb} 
\usepackage{amsmath} 
\usepackage{subfigure}
\usepackage{graphicx}
\usepackage{multirow}
\usepackage{multicol}
\usepackage{mathrsfs}
\usepackage{comment}
\usepackage{hyperref}

\usepackage{ulem}
\usepackage{color}

\newcommand{\revise}[2]{{#2}}

\title{Concise and Effective Network for 3D Human Modeling from Orthogonal Silhouettes}

\author{\hspace{-10pt}
Bin Liu$^{1,2}$, Xiuping Liu$^2$, Zhixin Yang$^3$, Charlie C.L. Wang$^4$\thanks{Corresponding Author; Email: changling.wang@manchester.ac.uk}
\\
\affiliation{
$^1$School of Mathematics and Information Science, Nanchang Hangkong University, Nanchang, China\\
$^2$School of Mathematical Sciences, Dalian University of Technology, Dalian, China\\
$^3$State Key Laboratory of Internet of Things for Smart City / Dept. of Electromechanical Engineering, \\University of Macau, Macau, China\\
$^4$Department of Mechanical, Aerospace and Civil Engineering, University of Manchester, United Kingdom
}
}

\begin{document}

\maketitle

\begin{abstract}
{
\it
In this paper, we revisit the problem of 3D human modeling from two orthogonal silhouettes of individuals (i.e., front and side views). Different from our prior work \cite{wang2003virtual}, a supervised learning approach based on \textit{convolutional neural network} (CNN) is investigated to solve the problem by establishing a mapping function that can effectively extract features from two silhouettes and fuse them into coefficients in the shape space of human bodies. A new CNN structure is proposed in our work to \revise{exact}{extract} not only the discriminative features of front and side views and also their mixed features for the mapping function. 3D human models with high accuracy are synthesized from coefficients generated by the mapping function.
Existing CNN approaches for 3D human modeling usually learn a large number of parameters (from 8.5M to 355.4M) from two binary images. Differently, we investigate a new network architecture and conduct the samples on silhouettes as input. As a consequence, more accurate models can be generated by our network with only 2.4M coefficients. The training of our network is conducted on samples obtained by augmenting a publicly accessible dataset.
Learning transfer by using datasets with a smaller number of scanned models is applied to our network to enable the function of generating results with gender-oriented (or geographical) patterns.
}
\end{abstract}



\section{Introduction}
In recent years, with the growth of demand in applications such as virtual try-on, customized design and body health monitoring, simple and effective human shape estimation techniques have caught more and more attention. Many approaches have been developed to generate 3D human models as mesh surfaces from photos as it is nowadays an easiest way to capture information by using smart phones. Considering the factor of a user-friendly interface, two images from two orthogonal views (i.e., the front and side views as conducted in \cite{wang2003virtual}) becomes an optimal solution if not the best.

With the help of a statistical model such as SCAPE~\cite{anguelov2005scape} that can capture the variation of human body's shape space, effort have been made to reconstruct 3D human models from silhouettes by minimizing the error between input silhouette images and the projected profiles of a parameterized 3D body shape \revise{ref.~}{}\cite{balan2007detailed,bualan2008naked,zhou2010parametric}. These methods in general need a complex matching algorithm to fit poses and shapes, which may result in a prohibitive complexity to prevent their usage in real-time applications.

To solve this problem, some researchers considered to establish the mapping between silhouettes and 3D models in a parametric representation in their shape space. They attempt to learn the mapping from handcrafted features by integrating linear~\cite{xi2007data,dibra2016shape} or non-linear~\cite{chen2009learning, boisvert2013three} projection. However, these approaches based on handcrafted features are not robust enough for input from real applications. Different from handcrafted features, \textit{convolutional neural network} (CNN) can make an end-to-end process to automatically extract features from a training dataset, which have embedded the application-oriented knowledge. Recently, researchers have made significant progress on CNN-based 3D human shape estimation from silhouette images \revise{ref.~}{}\cite{dibra2016hs,dibra2017human,ji2019human}. They use CNN to learn the non-linear mapping function between silhouettes images and the shape space of human bodies. The input silhouettes images represent the region with human body and the background as binary images, which contains a lot redundant information.
Directly learning features from binary images is less efficient as the meaningful information -- i.e., silhouettes are inherently sparse~\cite{2017Submanifold}.
To achieve an estimation with high accuracy, a large number of parameters will be trained in a time-consuming learning process; and it also results in a large memory consumption. Here we argue that only points on the silhouettes will be good enough to extract effective features to establish the mapping from orthogonal silhouettes to the shape space of human bodies in a concise network. \revise{}{A summary to compare the pros and cons of existing methods is given in Table~\ref{tab:introlimit}}.

\begin{table}[t]
\caption{\revise{}{Comparison of existing methods.}}
\label{tab:avegErrComparison} \vspace{5pt}
    \centering  \resizebox{\linewidth}{!}{
    \begin{tabular}{l|l}
   \hline
  Methods & Pros and cons \\
   \hline
    \hline
  \revise{}{Shape matching algorithms~\cite{balan2007detailed,bualan2008naked,zhou2010parametric}}  & \revise{}{Concise but time-consuming \& not robust} \\
    \hline
 \revise{}{Handcrafted features~\cite{xi2007data,dibra2016shape,chen2009learning, boisvert2013three}} & \revise{}{Efficient but not robust}  \\
   \hline
 \revise{}{CNN deep-learning~\cite{dibra2016hs,dibra2017human,ji2019human}} & \revise{}{Robust but with high memory consumption} \\
   \hline
   \end{tabular}}
    \label{tab:introlimit}
\end{table}

\revise{}{To overcome all existing problems}, we propose a novel CNN that is concise and effective. The network is trained on an augmented set of 3D human database, which is released by the authors of \cite{pishchulin2017building}. First, we construct a parameterized representation of 3D human bodies' shape space by applying \textit{principle component analysis} (PCA) to models normalized by their height. After that, each 3D human model can be compactly represented by the coefficients of the \revise{}{first} $k$ most important \textit{principle components} (PCs) instead of its original polygonal mesh. This can significantly reduce the difficulty of learning an effective network. Then, render-to-texture technique in OpenGL is used to generate the binary silhouette images from 3D models. The silhouette of every 3D human model is sampled into a fixed number of points, the $x$- and $y$-coordinates of which are employed as input for our CNN-based learning approach. The output of our network -- served as a non-linear mapping function is the coefficients of \revise{}{the first} $k$ PCs for the corresponding human model. A novel structure of CNN is proposed in our work, which can extract both the discriminative features from respective front and side silhouettes and the mixed feature to fuse the information from two views.
To generate results with geographical (or gender-oriented) patterns, we apply learning transfer to the above trained network by using datasets with a smaller number of scanned models.

In summary, our method shows the following merits:
\begin{itemize}
\item[$\bullet$] \textbf{Effectiveness:} Having two images taken from the front and side views as input, our network can generate 3D human models more accurate than other CNN-based methods~\cite{dibra2016hs,ji2019human}.

\item[$\bullet$] \textbf{Conciseness:} The number of parameters used in our network is at the size of 2.4M, which is much less than two other CNN-based methods for realizing the same function (i.e., 355.4M and 8.5M in \cite{dibra2016hs} and \cite{ji2019human} respectively).
\end{itemize}
Both are benefit from the novel network structure proposed in our method. Moreover, user study on a variety of individuals has conducted to verify the robustness of our approach. As a common merit of the end-to-end approaches, the computation for 3D human model estimation is very efficient, which allows it to be used in real-time applications. A mobile APP has been developed by using our network to produce customized clothes as an Industry 4.0 application.

\vspace{8pt} \noindent The rest of our paper is organized as follows. After reviewing the related work in Section \ref{secReview}, we present the technical approach of our CNN-based 3D human model generation in Section \ref{secTech}. The details of implementation are given in Section \ref{secImplementation}. Experimental results, comparisons and the user study on a variety of individuals are presented in Section \ref{secResult}. The application of customized design by using our approach will be presented in Section \ref{secApplication}. Lastly, our paper ends with the conclusion section.

\section{Related Work}\label{secReview}
Estimation 3D human shape from images is a popular research topic that has been widely studied for many years~\cite{balan2007detailed,chen2009learning,zhu2015predicting,kanazawa2018end,pavlakos2018learning}. Providing a comprehensive survey of all past researches has been beyond the scope of this paper. We only review the most relevant approaches that inspire this work.

\subsection{Human shape representation space}
Building a statistical model for human bodies has made significant improvement in recent years. The pioneer work presented in \cite{allen2003space} used PCA to model the representation space of human body shape without considering the human poses. With the help of this PCA-based statistical model, they could reconstruct a complete human surface from range scans based on template fitting. To obtain a better statistical model, researchers have tried to learn the shape variation space and pose variation space separately (e.g., the SCAPE~\cite{anguelov2005scape} and SMPL~\cite{loper2015smpl} models widely employed in many applications). The variation of human bodies is usually encoded in terms of deformation on triangular meshes with the same connectivity, which can be obtained by either template fitting (e.g., \cite{allen2003space,anguelov2005scape,wang2007volPara}) or the more advanced cross-parameterization technique (e.g., \cite{Kraevoy2004,Kwok2012a,Kwok2012b}). Some researchers intend to use the representation space from intrinsic features on mesh. Freifeld and Black~\cite{freifeld2012lie} adopted the principle geodesic analysis~\cite{fletcher2003statistics} to capture shape variation. Dibra et al.~\cite{dibra2017human} employed heat kernel signatures to obtain the shape representation space through convolutional neural network. In our approach, we adopted a dataset of human models with nearly unique pose to generate the shape space representation of human model by PCA-based statistical model. Skeleton-based representation in SMPL~\cite{loper2015smpl} can be used to further enrich our work by adding more poses into the training dataset.

\subsection{Shape estimation from images}
Estimating from images is a direct and low-cost mode for reconstructing shapes and poses of 3D human bodies, which has many applications \revise{ref.~}{}\cite{zhou2010parametric,Rogge2014Garment,neophytou2014layered}.
Some early methods (e.g.,~\cite{wang2003virtual,hilton2000whole,lee2000generating}) use a template model to minimize shape approximation errors, which heavily depends on the positions of feature point on 2D contours or the correspondence relationship between a 2D contour and the projected silhouette of a 3D model. These methods need to solve a complex corresponding problem, and therefore are not robust in general. Following the similar strategy, Zhou et al.~\cite{zhou2010parametric} presented an image retouching technique for realistic reshaping of human bodies in a single image. Zhu et al.~\cite{zhu2015predicting} integrated both image-based and example-based modelling techniques to create human models for individual customers based on two orthogonal-view photos. Song et al.~\cite{song20163d} selected critical vertices on 3D registered human bodies as landmarks. After that, they learned a regression function from these vertices to human shape representation space defined by SCAPE~\cite{anguelov2005scape}. The time-consuming error minimization process is involved in these approaches. Unlike our end-to-end approach, they are not efficient enough to be used in real-time applications.

Researchers also developed techniques to estimate 3D human shape from a single or a few silhouettes in a standard posture by handcrafted features. A probabilistic framework representing the shape variation of human bodies was proposed by Chen et al.~\cite{chen2010inferring} to predict 3D human shape from a single silhouette. Boisvert et al.~\cite{boisvert2013three} used geometric and statistical priors to reconstruct the human shape
form a frontal and a lateral silhouettes. Differently, Alldieck et al.~\cite{alldieck2018video} employed a video to estimate the shape and texture by transforming the silhouette cones corresponding to dynamic human silhouettes. The major problem of these approaches based on handcrafted features is that they have prohibitive time complexity to be used in practical applications.

To resolve the efficiency problem, more and more research effort has been paid to develop an end-to-end approach with the help of CNN techniques. Toward this trend, Dibra et al.~\cite{dibra2016hs} and Ji et al.~\cite{ji2019human} reconstructed 3D human shape from two orthogonal silhouette images based on a modified AlexNet~\cite{krizhevsky2012imagenet} and multiple dense blocks~\cite{huang2017densely}. The major difference between these two approaches is their training mode. Ji et al.~\cite{ji2019human} first learned two networks from the front and the side views separately, and then trained a new network from the features learned in the previously determined two networks. To be less influenced by the poses of human bodies, Dibra et al.~\cite{dibra2017human} conducted multi-view learning on a quite complex cross-modal neural network, which cannot be trained simultaneously. A common problem of these methods~\cite{dibra2016hs,dibra2017human,ji2019human} is that direct learning from binary images by convolution operators is inefficient as the silhouette features are very sparsely represented in the binary images. More discussion about the learning efficiency influenced by the sparsity of information can be found in \cite{2017Submanifold}.

Recently, researchers have developed a variety of techniques for estimating the 3D human shape from a single RGB image with the help of CNN(e.g., \cite{2018Learning,2018End,Huang2018DVV,2019PIFu,Zheng2019DeepHuman}). However, the accuracy of resultant models generated by these approaches is in general worse than the results obtained from orthogonal silhouettes because of the absence of other views. Moreover, when photos with tight clothes are used as input, the privacy of a user is better protected by silhouette images.

\subsection{Convolutional neural network}
In many computer vision tasks such as image classification, recognition and segmentation, it is an important role to obtain highly discriminative image descriptors. For this purpose, a wide range of neural networks have been proposed -- such as AlexNet~\cite{krizhevsky2012imagenet},
VGGNet~\cite{simonyan2014very}, GoogleLeNet~\cite{szegedy2015going}, U-net~\cite{ronneberger2015u}, ResNet~\cite{he2016deep} and DenseNet~\cite{huang2017densely}. These prior researches have introduced some frequently used operations in CNN architecture, such as convolution layer, pooling layer and activation layer. In addition, they also provided inspirational ideas on other aspects of network design. For example, dropout layer is used to prevent over-fitting,
batch normalization is utilized to accelerate the convergence of network
and identify block is proposed to enhance feature reuse. All these concepts inspire the architecture design of our network.

As the input of our network is a well-organized set of points, the approach with similar well-organized information such as TextCNN~\cite{kim2014convolutional} and PointNet~\cite{qi2017pointnet} also provide inspiration to the design of our network.
TextCNN has applied in sentence-level classification tasks and performed convolution on top of word vectors.
PointNet was used to deal with point sets and it also provided an architecture for applications ranging from 3D shape classification to semantic segmentation.
In their approaches, filters are first used to extract features from input datasets, which is followed by applying the maximal pooling operation over the feature map to capture the global descriptors. In final steps, fully connected layers are employed to obtain the labels of input. We use a similar routine to design the architecture of our network. However, to capture more local structural information of sample points on silhouettes, maximal pooling operations are only performed in the local region of each feature map in our approach.

\section{Technique Approach}\label{secTech}
In this section, we first introduce the PCA-based shape representation that spans the space of 3D human models. Then, the method of data augmentation is presented to enrich the training dataset. After these steps of preparation, the structure and the loss function of our network are presented.

\begin{figure}
\centering
\includegraphics[width = \linewidth]{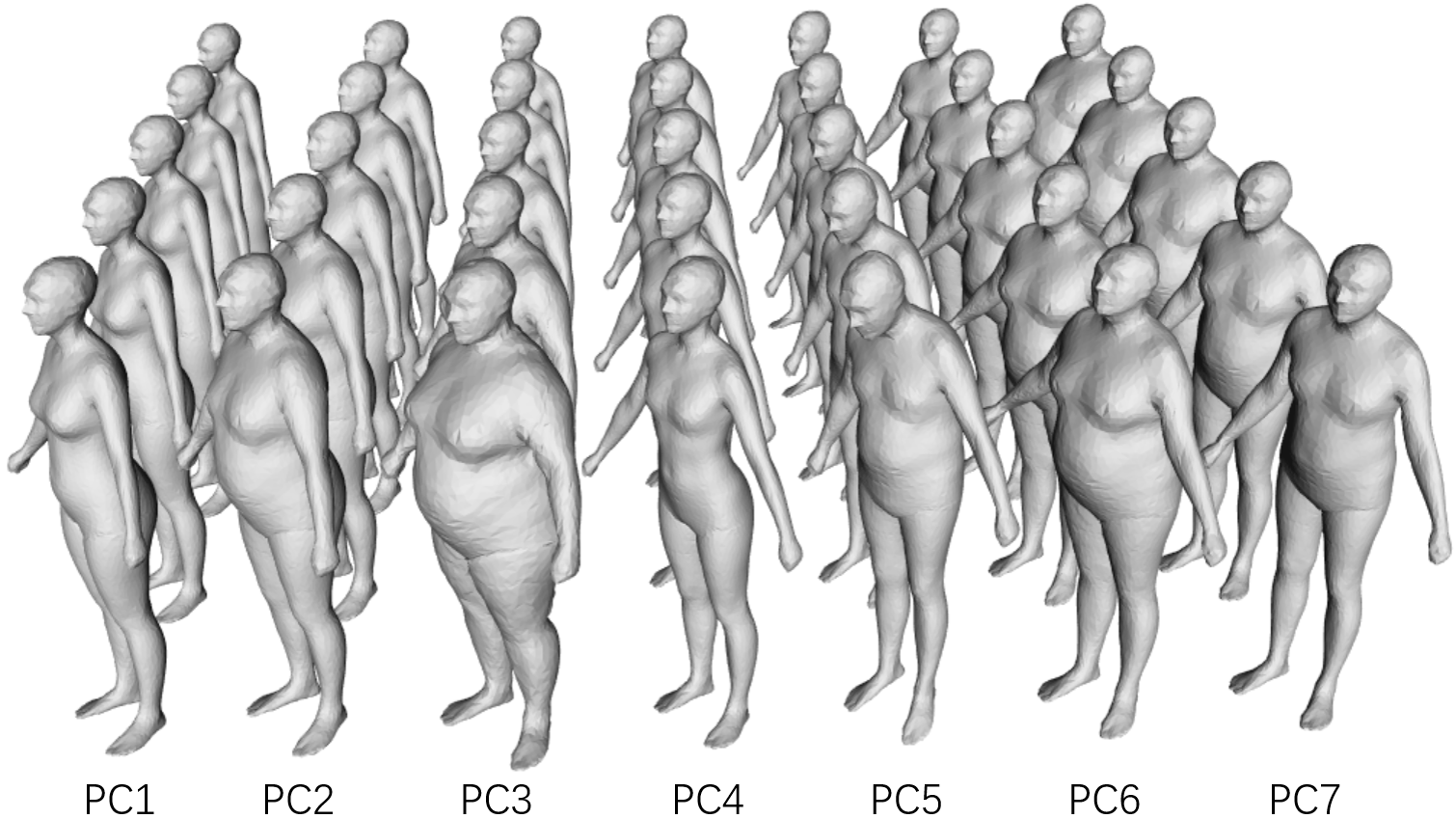}
\caption{In our CNN-based approach, 22 \textit{principal components} (PCs) are employed to span the shape space of 3D human models ($k=22$). The first seven PCs are displayed to visualize the shape variations.
}\label{fig:pcdata}
\end{figure}

\subsection{Shape space representation}
\textit{Principal component analysis} (PCA) has been widely used to construct statistical models for representing the shape variation of 3D human bodies~\cite{anguelov2005scape,allen2003space,chu2010statisticHuman,wang2012geometric,pishchulin2017building}, where all models have the same mesh connectivity. We adopt the strategy of normalizing the height of all models before PCA \cite{kwok2014volHuman} to better capture the shape variation. In our approach, we represent the shape of each human body by a vector $\varphi_s \in \mathbb{R}^k$ as
\begin{equation}\label{Eq:pca}
    B_s = B(\varphi_s) = \bar{B} + \Omega \varphi_s
\end{equation}
where $B_s \in \mathbb{R}^{3N}$ denotes a body shape with $N$ vertices,
$\bar{B}$ represents the average body shape of all models in the dataset and $\Omega \in \mathbb{R}^{3N \times k}$ is a linear space formed by $k$ principal component vectors. In our implementation, $k$ is set as the minimum value when the statistic model based on PCA can capture more than 97\% of the original shape. The shape of each human model is then represented by the vector $\varphi_s$ with $k$ components -- named as \textit{shape coefficients}. This representation is much more compact than the original mesh representation with $3N$ coefficients for the $N$ vertices. When the poses of all models in a dataset are the same, PCA can mainly capture the shape variation in bodies and not be influenced by the pose variation. We use the dataset of $4,308$ individuals released by the authors of \cite{pishchulin2017building} to build the space of human shape representation, where all models are in a neutral pose and are generated by fitting a template mesh to the CAESAR dataset~\cite{robinette2002civilian}.
In detail, we first select $4,306$ models from this database -- two are left out as they are incomplete and highly distorted.
Then, these models are split into two dataset with a ratio of $4:1$, where the set with $3,444$ models are used to build a training dataset (will be further augmented below) and the set with $862$ models are employed as the test dataset. PCA model is build on the set with $3,444$ models. An illustration of the first $7$ principal components can be seen in Fig.~\ref{fig:pcdata}.

\subsection{Training data augmentation}\label{subsecAugmentation}
In general, the set of training data should be large enough for machine learning and it is expected to be uniformly distributed to cover the whole space -- especially for the task of regression \revise{ref.~}{}\cite{lin2017focal}. For this purpose, the training set is further augmented by inserting more samples according to a metric defined on the Euclidean distances between the vectors of shape coefficients.
We aims at enlarging the dataset meanwhile improving its distribution to become more uniform.

\begin{figure}[t]
\centering
\subfigure[Our initial dataset with $3,444$ samples]
{\includegraphics[width = .9\linewidth]{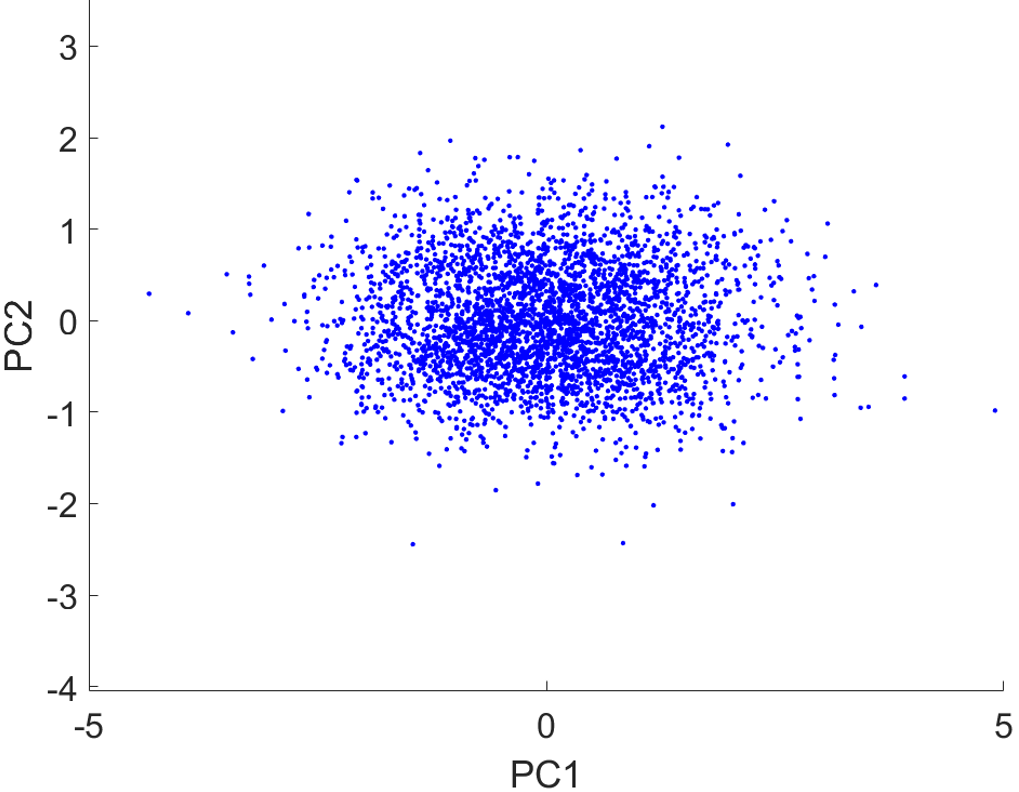}}
\subfigure[The augmented dataset with $8,941$ samples]
{\includegraphics[width = .92\linewidth]{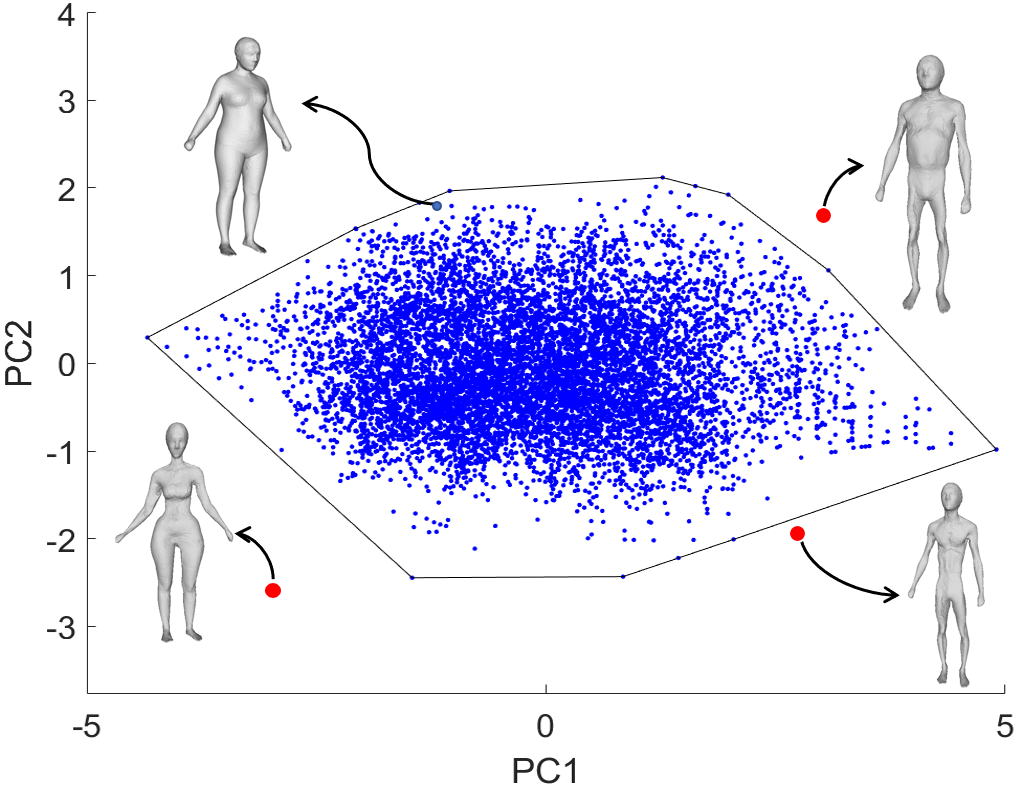}}
\caption{An illustration of sample distribution on PC1 and PC2. After applying our refinement based data augmentation, the distribution has become more uniform and the newly added samples are all inside the convex-hull of the original samples. Sampling from the Gaussian distribution along each PC may generate a model with unrealistic body shape (see the three examples in (b) as red points) that are located outside the convex-hull of original data points.
}\label{fig:augment}
\end{figure}

During data augmentation, we use the K-nearest neighbor search to obtain the neighboring set $\mathcal{N}_s$ for each sample $\varphi_s$ in the dataset. Here $K=11$ is adopted in our experiments. We define the \textit{neighboring radius} of a sample $\varphi_s$ as
\begin{equation}\label{Eq:dis}
r_s = \frac{1}{K}\sum\limits_{\varphi_i \in \mathcal{N}_s}{\|\varphi_i - \varphi_s\|}.
\end{equation}
The minimal neighboring radius $r_{min}$ among all samples can be calculated by $r_{min} = \min \{r_s\}$. Then, the following refinement step is iteratively applied: for any pair of neighboring samples $\varphi_s$ and $\varphi_q$, if $\|\varphi_s - \varphi_q\| > 1.8 r_{min}$, we insert a new sample in the middle as $\varphi_p = (\varphi_s + \varphi_q) / 2$.
The iteration is repeated two or three times.
After refinement, we obtain an augmented dataset of human models with $8,941$ samples (see Fig.\ref{fig:augment}).

Here we do not employ the method of sampling from the Gaussian distribution along each PC, which is used in other papers~\cite{dibra2016hs,dibra2017human,ji2019human}. This is because that simply generating a new sample from Gaussian distribution may lead to unrealistic body shapes (see the samples shown in Fig.\ref{fig:augment}(b)). Our method of data augmentation can effectively prevent the generation of unrealistic samples.

\begin{figure*}
    \centering
    \includegraphics[width = .9\linewidth]{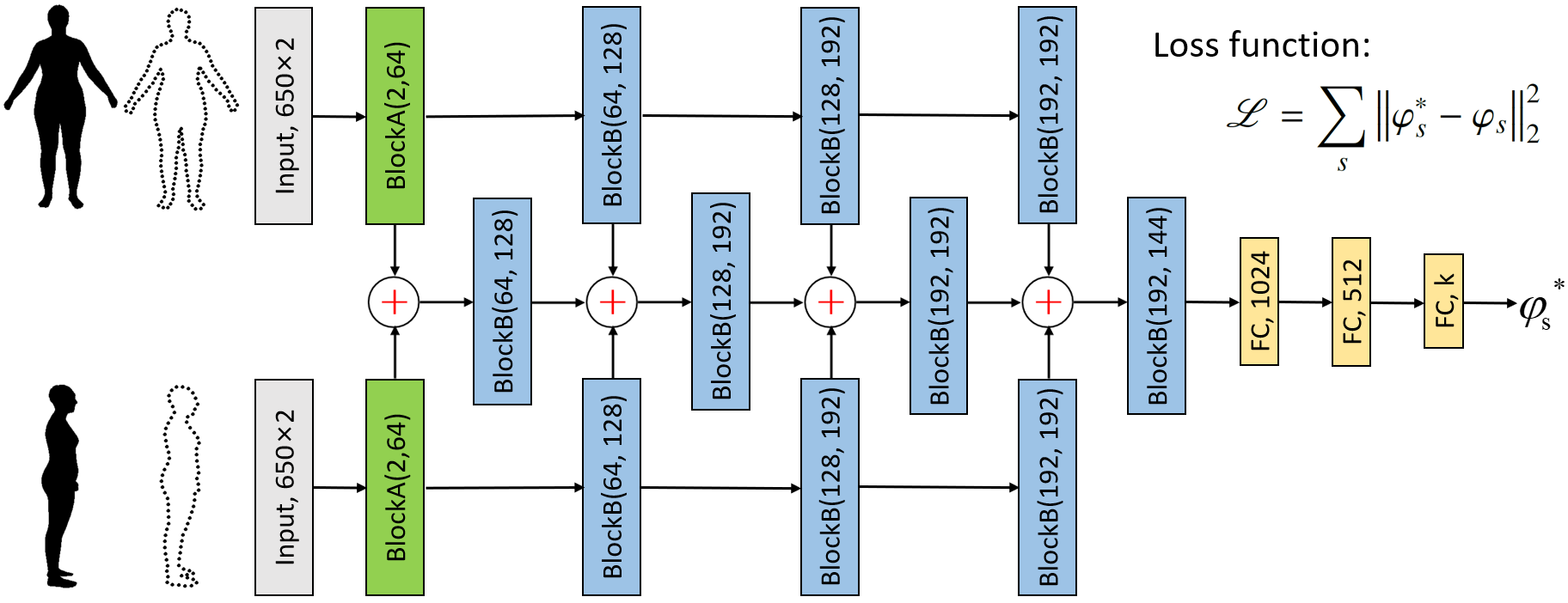}
\caption{The overview of our CNN's architecture, \revise{}{which takes the two orthogonal silhouettes as input and outputs the estimated shape coefficient.} \revise{where two types of `Blocks' are employed.}{It employs two types of `Blocks',}
each `Block' contains 3 or 4 operations, and the details can be found in Fig.~\ref{fig:block}.
'+' represents the additive operation between output layers of `Blocks', which facilitates the information communication between `Blocks' of different views.
FC denotes a fully connected layer and the following number gives the number of hidden neurons. $\varphi_{s}^*$ represents the output of network and it is supervised by the shape coefficient $\varphi_{s}$ of ground truth.}
\label{fig:architecture}
\end{figure*}

\begin{figure}
\centering
\includegraphics[width = \linewidth]{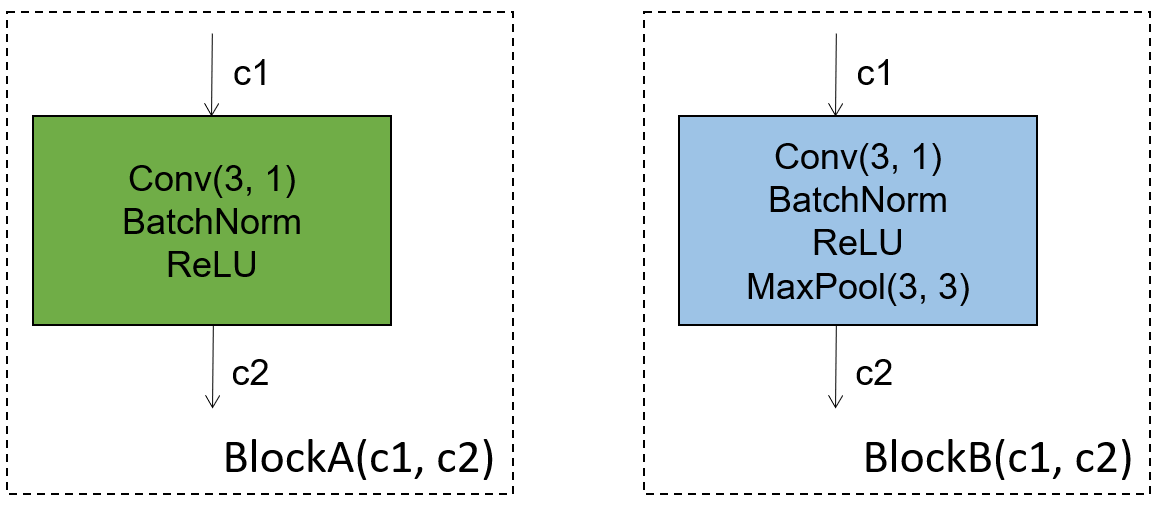}
\caption{{The structures of BlockA(c1, c2) and BlockB(c1, c2), where c1 and c2 represent the input and output number of components in feature maps. BatchNorm means the batch normalization operation and ReLU is the activation function.}}\label{fig:block}
\end{figure}

\subsection{Silhouette sampling}
Now we introduce how to generate the silhouette images of a 3D human model and the sampling points on the silhouette contours.
For each sample $\varphi_s$, we reconstruct its 3D mesh model by Eq.(\ref{Eq:pca}). Then, the 3D model is rendered in a window with $500\times600$ pixels as binary images along its front-view and side-view (see the most-left of Fig.\ref{fig:architecture}). In order to simulate the real scenario of taking photos for individuals, we conduct perspective projection to render models and also add a Gaussian perturbation to the position of camera (i.e., setting the mean position at the center of image and the standard deviation of $0.05$m with the image height as $2$m). The contour extraction technology \cite{suzuki1985topological} is applied to obtain the closed boundary from each silhouette image. Then, the contour is uniformly sampled into $M$ ordered points, the coordinate of which are transformed back to the scale of unit body height. Origin of all sample points are located at their average position.

Every list of ordered sample points starts from a highest point located at the top and center of the model's head. All points are sorted in the anti-clockwise order. Moreover, a copy of the ending point and a copy of the starting point are inserted at the beginning and the end of the list respectively to solve the problem of convolution at the boundary of input. The number of sample points will have influence on the accuracy of 3D shape estimation. How to choose the number of samples will be discussed later in Section \ref{secImplementation}.

\subsection{Network architecture for learning}\label{subsecNetwork}
Front-view and side-view can adequately convey the main shape of a human body from different aspects. Their boundary points also encode information with local and global structures. A good network architecture is important for the effectiveness of learning a mapping function between contour sampling points and the shape coefficients.
\textit{Convolutional neural networks} (CNNs) are well known for their capability of automatic feature extraction and adaptability of highly nonlinear mapping. We propose a novel CNN architecture for the problem of 3D human shape estimation from orthogonal silhouettes. Our network will be trained to extract features from the local and global structures, which will be mapped to the shape coefficients for generating 3D human model from PCs.

There are two existing CNN architectures for solving the problem of 3D human modeling from orthogonal silhouettes. Both use binary images as input.
\begin{itemize}
\item[$\bullet$] Dibra et al.~\cite{dibra2016hs} proposed an architecture which is similar to AlexNet~\cite{krizhevsky2012imagenet} to learn the mapping function. They first stack the front-view and the right-view images to form two channel images considered as the input of their network. The network employed in this approach is very dense.

\item[$\bullet$] Ji et al.~\cite{ji2019human} addressed the same problem by an an architecture that contains three pipelines. Two of them are used to extract features from the front-view and the right-view silhouette images. The third one is used to concatenate the features learned from two separated pipelines to be further processed by fully connected layers.
\end{itemize}
Differently, we use samples on silhouette contours as input. A novel network is proposed with the architecture as illustrated in Figs.~\ref{fig:architecture} and \ref{fig:block}. Our network can extract discriminative features from silhouette sample points of two orthogonal views and also fuse the features learned from different views after each block of learning. The additional blocks in the middle pipeline facilitate information communication between two views and help to fuse the local structures of human body in two views. Moreover, different from \cite{ji2019human} that three pipelines are trained one by one, the three pipelines in our architecture can still be trained together in an end-to-end mode. This is easier for implementation. In order to extract local and global structure information from sample points of silhouette contours, the following three strategies are employed in our network design which facilitate feature propagation, feature reuse and feature aggregation.
\begin{itemize}
\item[$\bullet$] Convolution layer with size 3 and step 1 (Conv(3,1)) is applied to perceive local structures on the silhouette contour.  Furthermore, multiple such layers are employed to enlarge the receptive field of convolution.

\item[$\bullet$] The layer of maximal pooling with size 3 and step 3 (MaxPool(3,3)) is adopted to aggregates local features to global features step by step. In our network architecture, the maximal pooling operation is applied 4 times to reduce the size of feature maps into $(1/3)^4=1/81$ of its original size $648 \times 1$.

\item[$\bullet$] Information communication between blocks of two views can help to detect more coherent local structures of human bodies. Note that the additive operation is employed in our fusion pipeline. The benefit is twofold. While it can make the features from different views fused together, it can also largely decrease the number of parameters comparing to the commonly used concatenation operator for fusion.
\end{itemize}
All fully connected layers use the ReLU activation function except the final layer. In short, we intend to provide a concise and effective solution, which takes full advantage of samples from two orthogonal silhouette contours.

Unlike TextNet~\cite{kim2014convolutional} and PointNet~\cite{qi2017pointnet}, we do not conduct a final maximal pooling operation applied to the whole feature map in our network. This is because that such an operation will discard too much local information. It is very useful for a classification task but not a regression task as what we propose in this paper. Therefore, we only embrace the maximal pooling operation in a region with size restriction, which can effectively extract and use the information of local structures. A comparison will be given in Section \ref{subsecRegionalPooling} to further verify the effectiveness of this choice (see Fig.~\ref{fig:fusion_pooling}).

\subsection{Loss function}
We now discuss the loss function used in training. For every 3D model in the training dataset represented as $\varphi_s$ in the human shape space with the predicted shape coefficients as $\varphi_s^*$, the error metric for shape difference between the original 3D model and the predicted 3D model can be defined as the sum of squared vertex-to-vertex distances. The total error on all models is then used as the loss function as
\begin{equation}\label{Eq:loss_function_vertex}
    \mathscr{L} = \sum_{s} {\left\|B(\varphi_s^*) - B(\varphi_s) \right\|}_2^2
\end{equation}
with $B(\cdot)$ defined in Eq.(\ref{Eq:pca}). The error for each model can then be re-written as
\begin{equation}\label{Eq:loss_function_derive}
\begin{split}
\|B(\varphi_s^*) - B(\varphi_s)\|_2^2 & = \| (\bar{B}+\Omega \varphi_s^*) - (\bar{B}+\Omega \varphi_s) \|^2_2 \\
 & = \|\Omega(\varphi_s^* - \varphi_s)\|^2_2 \\
 & = (\varphi_s^* - \varphi_s)^T\Omega^T\Omega(\varphi_s^* - \varphi_s)
\end{split}
\end{equation}
When using orthonormal vectors to form the shape space $\Omega$ (i.e., the PCs obtained by PCA), we can have $\Omega^T \Omega = I$. As a consequence, the loss function can be simplified to
\begin{equation}\label{Eq:loss_function}
    \mathscr{L} =  \sum_{s} {\left\|\varphi_s^* - \varphi_s \right\|}_2^2.
\end{equation}
This is the loss function used in our training process. Comparing to Eq.(\ref{Eq:loss_function_vertex}), the evaluation of loss function defined in Eq.(\ref{Eq:loss_function}) can save a lot of memory and computing time.

\section{Metrics Evaluation and Implementation Details}\label{secImplementation}
We have implemented the proposed network in PyTorch, and the source code is available to public\footnote{\url{https://github.com/liubindlut/SilhouettesbasedHSE}}. Our network is trained with all parameters randomly initialized, and we train the network by using the Adam optimization algorithm \cite{kingma2014adam} running on a PC with an Intel(R) Core(Tm) i7-8700 CPU $@$ 3.2GHz, 16GB of RAM and a GeForce GTX 2080 GPU. The epochs and batch size are set as 500 and 128 respectively with the learning rate {$1.0\times 10^{-5}$}, and we use the default values for all other parameters.

In this section, we will discuss a few detail decisions about our approach, including the number of sampling points on silhouette contours, the functionality of fusion blocks and the usage of regional max pooling. To facilitate the discussion of these decisions, the metrics used for evaluation the performance of a network are first introduced below.

\begin{figure}[t]
\centering
\includegraphics[width = \linewidth]{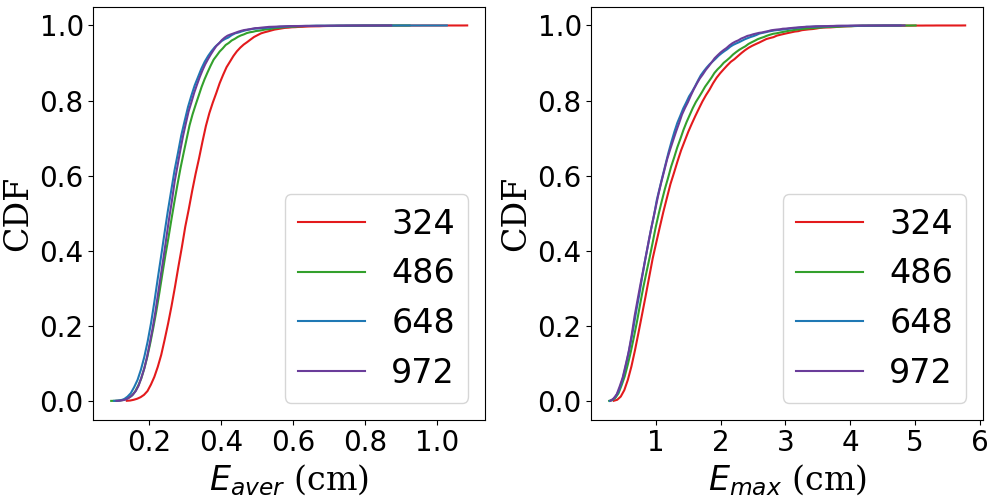}
\caption{Comparing the performance of our network when using different numbers of sampling points on silhouette contours. The CDF curves are generated on the training dataset to tune the number of sample points used for each silhouette.}\label{fig:sampling_num}
\end{figure}

\begin{table}
\caption{{Statistic by using different numbers of contour points}}
\label{tab:ContourPntNum} \vspace{5pt}
    \centering \small
\begin{tabular}{r|c|c|c|c}
\hline
\# of Pnts. & $324$ & $486$ & $648$ & $972$ \\
\hline
\hline
Coefficients \# in Network & 1.84M & 2.14M & 2.43M & 3.02M \\
Average Comp. Time (ms) & $33.3$ & $33.8$ & $34.3$ & $35.2$ \\
\hline
\end{tabular}
\end{table}

\begin{figure}[t]
\centering
\includegraphics[width = \linewidth]{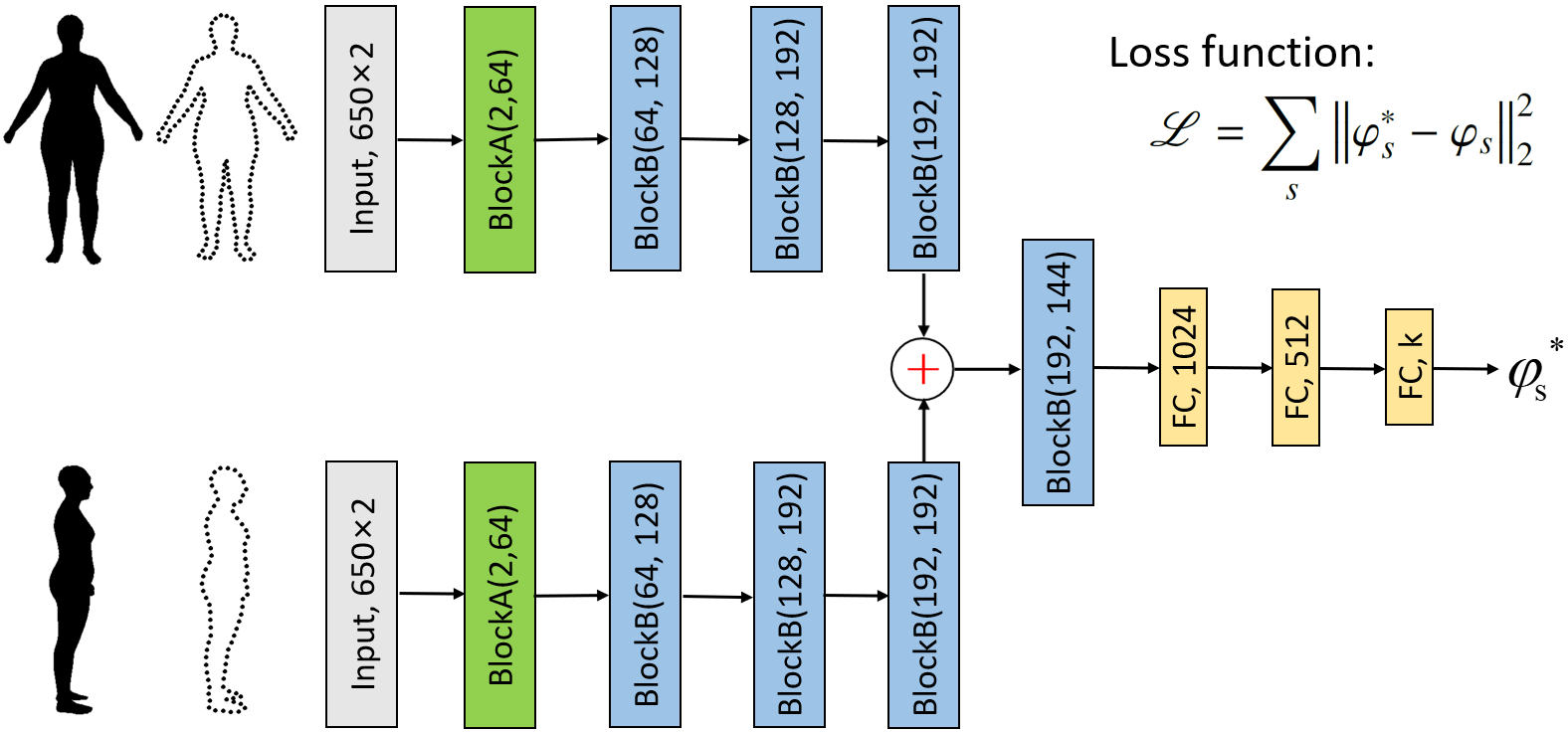}
\caption{{The architecture of our network after removing the fusion blocks.}}
\label{fig:withoutfusion}
\end{figure}
\begin{figure}[h]
\centering
\includegraphics[width = \linewidth]{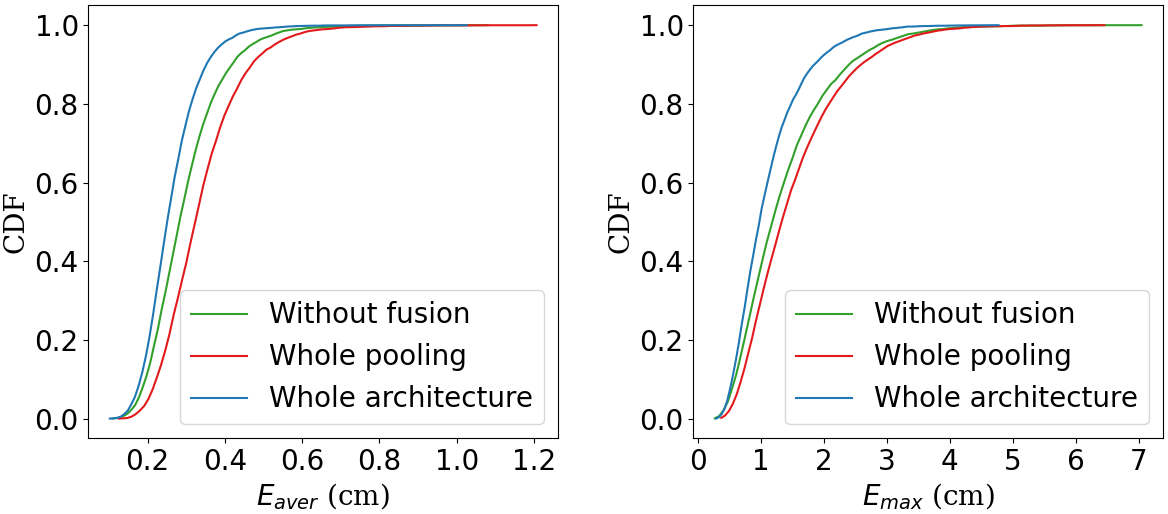}
\caption{{Comparison of CDF curves generated on the training dataset by using different strategies of network -- 1) with vs. without fusion blocks and 2) regional or whole-map max pooling.}
}\label{fig:fusion_pooling}
\end{figure}
\begin{figure}[h]
\centering
\includegraphics[width = .9\linewidth]{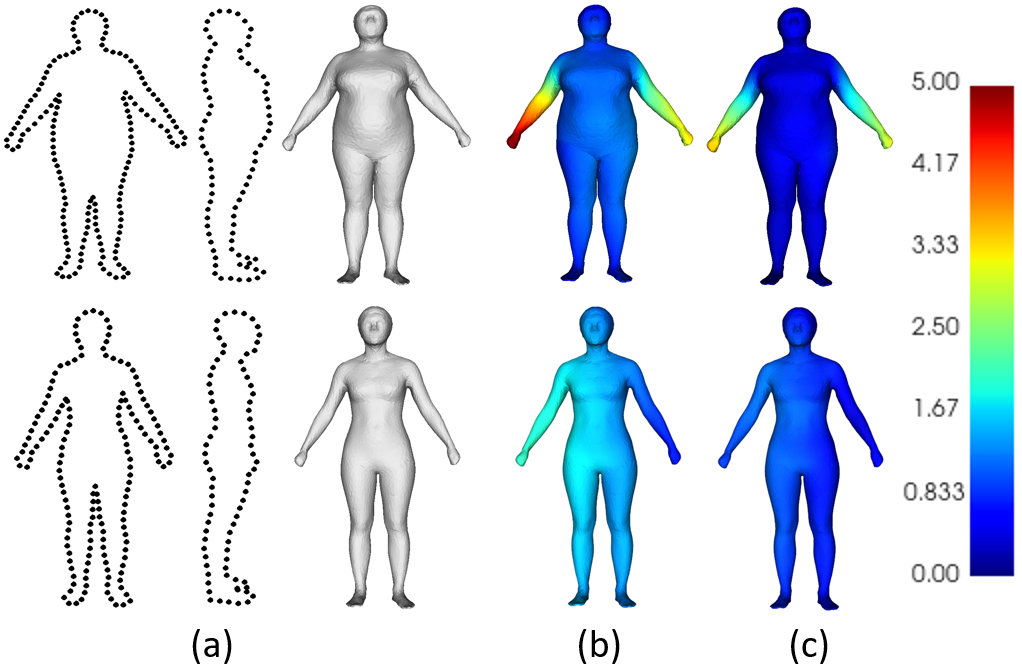}
\caption{{Comparison of reconstruction results to illustrate the effectiveness of fusion blocks -- (a) the sample points on original silhouette and the ground-truth model, (b) the results generated by a network without fusion blocks (i.e., the architecture in Fig.\ref{fig:withoutfusion}) and (c) the results generated by our network with fusion blocks. It is easy to find from the colorful error maps given in (b) and (c) that the reconstruction errors can be significantly reduced after adding the fusion blocks. These two example are from the training dataset.}
}\label{fig:fusion_examples}
\end{figure}

\begin{figure}
\centering
\includegraphics[width = \linewidth]{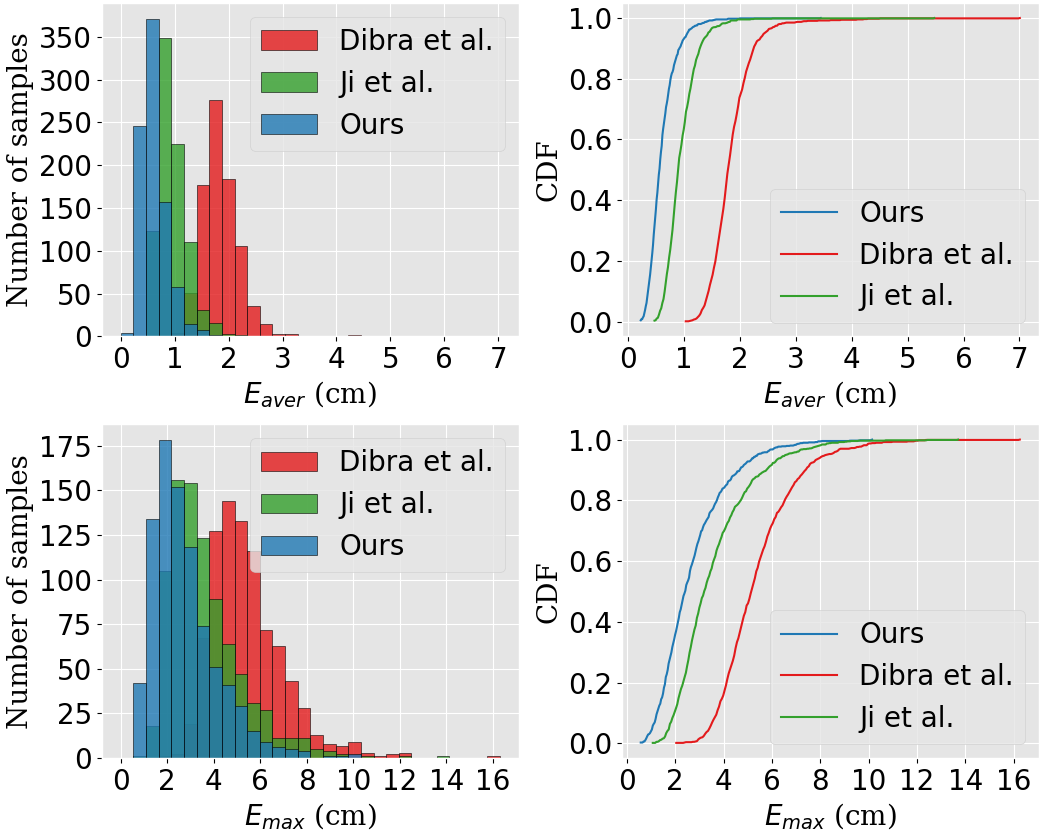}
\caption{{The distributions of geometric errors on test samples. The unit of errors is centimeter (CM). Left shows the histogram of frequency distribution and the corresponding probability cumulative distribution functions (CDFs) are shown in the right.}}\label{fig:datasetsDistribution}
\end{figure}

\subsection{Metrics for evaluation}
We employ two shape-based metrics to provide the quantitative evaluation for the accuracy of reconstructed 3D models. For all samples in the test dataset, they have the ground-truth surfaces represented as triangular meshes. From an estimated vector of shape coefficients $\varphi_s$, the estimated 3D shape can be obtained as a triangular mesh with the same connectivity. Therefore, we can use the maximal error $E_{max}$ and the average error $E_{aver}$ of vertex-to-vertex distances to measure the deviation between the reconstructed and the ground-truth surfaces. Specifically, these two metrics can be defined as
\begin{equation}\label{Eq:max1}
E_{max} = \max_{i = 1, \ldots, N} \left\{ \| \mathbf{v}_i - \mathbf{v}_i^* \|_2 \right\}
\end{equation}
and
\begin{equation}\label{Eq:aver1}
E_{aver} = \frac{1}{N} \sum_{i=1}^N{\left\| \mathbf{v}_i - \mathbf{v}_i^* \right\|_2},
\end{equation}
where $\mathbf{v}_i$ and $\mathbf{v}_i^*$ are the positions of the $i$-th vertex on the ground-truth and the reconstructed meshes respectively,
and $N$ is the total number of vertices on the mesh. With the help of these two metrics, we can use the histogram of frequency distribution and the \textit{cumulative distribution function} (CDF) to measure the performance of each algorithm on a whole dataset.
In addition, we can also make a rough estimation by the average values among $H$ samples in a dataset as
\begin{equation}
\bar{E}_{max} =\frac{1}{H}\sum_{j=1}^H E_{max}^j,
\quad
\bar{E}_{aver} = \frac{1}{H}\sum_{j=1}^H E_{aver}^j
\end{equation}
with $E_{max}^j$ and $E_{aver}^j$ being the corresponding errors of the $j$-th sample,  $H$ is the number of dataset.

As all human models generated by network having the same mesh. Feature curves for measuring intrinsic metrics can be predefined on the mesh surface and mapped to newly generate human models by barycentric coordinates. Besides of the above geometric metrics, we also follow the strategy of Dibra et al.~\cite{dibra2016hs} by evaluating intrinsic metrics on human bodies. Specifically, the errors on three girths (B:Chest, W:Waist and H:Hip) are evaluated in our experiments.

\subsection{Number of contour points}
Given silhouette images in a fixed resolution, the number of points employed to sample the silhouette contours will have influence on the capability to reconstruct an accurate 3D human model. Here we tune this parameter by studying the accuracy of reconstruction -- i.e., $324$, $486$, $648$ and $972$ contour points are tried respectively. CDFs for both $E_{max}$ and $E_{aver}$ are generated as shown in Fig.\ref{fig:sampling_num}. It is easy to find from the figure that the network trained by using more contour points will have more samples with errors less than a given threshold for both $E_{max}$ and $E_{aver}$. In other words, the network trained from more contour points is more accurate. No lunch is free. Training by using more contour points will lead to higher memory cost and therefore also longer computing time in reconstruction. Specifically, the resultant numbers of coefficients for storing a trained network and the average computing time for reconstructing a 3D human model are given in Table \ref{tab:ContourPntNum}. We choose $648$ contour points in practice by considering the balance between computational cost and accuracy.

\subsection{With or without fusion blocks}
Now we study the functionality of the blocks used in our network for fusing features between front and side views. After removing all the blocks for fusion, our network will degenerate into another architecture as shown in Fig.\ref{fig:withoutfusion}, which is similar to the network presented in \cite{ji2019human}. The training dataset is then employed to check the effectiveness of these two networks. The curves of CDF are generated for both $E_{max}$ and $E_{aver}$ as shown in Fig.\ref{fig:fusion_pooling}. It can be observed that the accuracy of reconstruction can be significantly improved after adding the fusion blocks. The values of $\bar{E}_{aver}$ and $\bar{E}_{max}$ drop from $0.31$ to $0.27$ and from $1.44$ to $1.15$ respectively. The comparison of reconstruction with vs. without fusion blocks on two examples can also be found in Fig.\ref{fig:fusion_examples}, where the shape deviation errors are visualized as color maps. \revise{}{Based on the results obtained from experiment, we believe that our architecture can be used as a basic network, which can derive different variants by modifying the structures of BlockA and BlockB according to the strategy of multi-modal fusion learning~\cite{gao2020survey}.}

\subsection{Regional or whole map pooling}\label{subsecRegionalPooling}
After verifying the functionality of fusion blocks, we study the effectiveness of using regional max pooling in our network design. As discussed above, when applying maximal pooling operation to the whole feature map, the network will discard too much local features which however are important to our human model reconstruction problem. We compare the results by using regional maximal pooling (as MaxPool(3,3) in our BlockB design -- see Fig.\ref{fig:block}) with the results by using whole-map max pooling operation. From the curves of CDF given in Fig.\ref{fig:fusion_pooling}, it is found that our design by using regional pooling gives more accurate reconstruction.

\begin{figure*}[t]
\centering
\includegraphics[width = .95\linewidth]{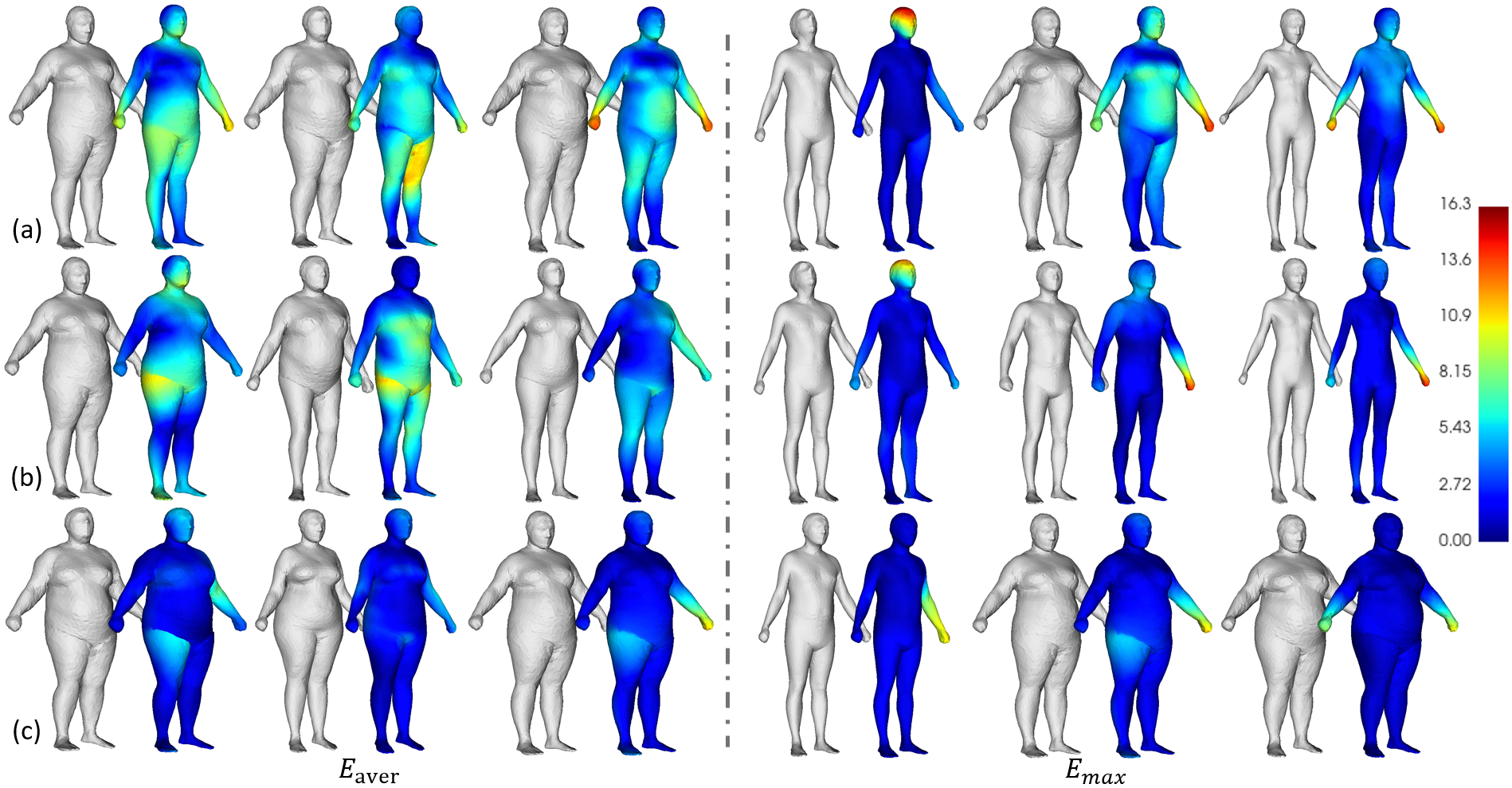}
\caption{
The illustration of three worst reconstruction with the metric $E_{aver}$ (left) and the metric $E_{max}$ (right) among all the samples in the test dataset for different methods -- (a) the results of Dibra et al.~\cite{dibra2016hs}, (b) the results of Ji et al.~\cite{ji2019human} and (c) our results. For every example, a pair of models are displayed where the ground-truth models are shown in gray color and the colorful maps of reconstructed models are employed to visualize the distribution of point-to-point distance errors in the unit of centimeter.
}\label{fig:examples}
\end{figure*}


\section{Results}\label{secResult}
The work presented in this paper focuses on reconstructing 3D human models from two orthogonal silhouette images by using the deep learning technique. In this section, we compare the results generated by our approach with the state-of-the-art methods~\cite{dibra2016hs,ji2019human} in different aspects, including the number of parameters, the geometric accuracy of reconstruction and the errors on intrinsic measurements. For conducting fair comparisons, we move the centers of silhouette in both views into the center of binary images as the input of their methods. Moreover, user study is conducted on a variety of individuals to verify the performance and robustness of our approach.

\subsection{Number of parameters}
Dibra et al.~\cite{dibra2016hs} used the architecture of AlexNet~\cite{krizhevsky2012imagenet} to estimate the shape parameters. Considering the limited memory of GPU, we choose the implementation that takes two silhouette images as two channels introduced in their paper. They tested silhouette images with resolution of $192 \times 264$, and the number of their parameters is about 355.4M. Ji et al.~\cite{ji2019human} employed the structure of DenseNet~\cite{huang2017densely} in their architecture, which results in a network with 8.5M parameters when using the input images at the resolution of $64 \times 64$. Moreover, they also discussed the influence of image resolution in their paper and claim that results with similar accuracy will be obtain even after doubling the resolution of input images. Differently, our network needs to store only 2.4M parameters. In practical applications, the memory consumption of a network is a very important cost to be controlled as the AI-engine is always operated on the cloud servers. In terms of demanded storage, our approach needs only 10M while the other two methods need about 1.4G and 34M storage space respectively. The superiority of silhouette and accuracy of reconstruction are discussed below.

\begin{table}
\caption{Quantitative comparisons for different methods on test samples (Unit: centimeter). The best method is highlighted.}
\label{tab:avegErrComparison} \vspace{5pt}
    \centering \small
    \begin{tabular}{c|c|c|c}
   \hline
  & Dibra et al.~\cite{dibra2016hs} & Ji et al.~\cite{ji2019human} & Our Approach \\
   \hline
    \hline
  {$\bar{E}_{aver}$} & 1.87          & 0.97          & \textbf{0.62} \\
    \hline
   {$\bar{E}_{max}$}  & 5.46       & 3.63           & \textbf{2.76} \\
   \hline
   \end{tabular}
    \label{tab:metnods}
\end{table}
\subsection{Accuracy of reconstruction}
We now evaluate the accuracy of 3D human models reconstructed by our method and compare to the state-of-the-art~\cite{dibra2016hs,ji2019human}. Both histograms and colorful error maps are employed to visualize the errors.

\begin{figure}
\centering
\includegraphics[width =\linewidth]{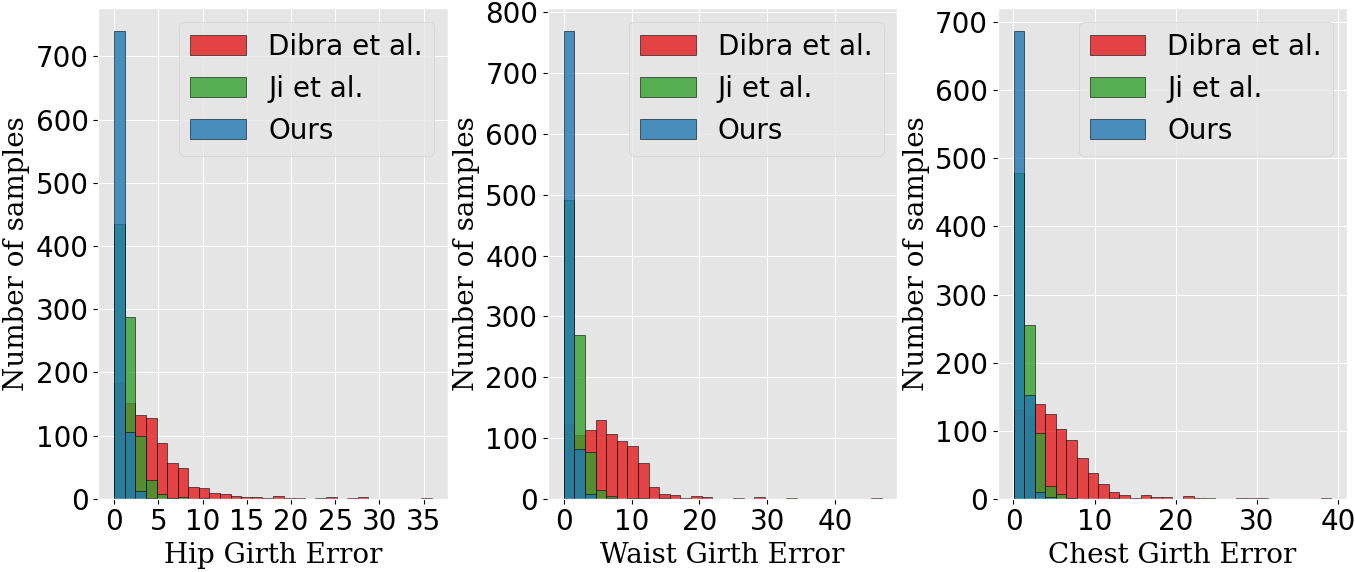}
\caption{The distributions of girth errors on test samples (Unit: centimeter).}\label{fig:GirthErrorDistribution}
\end{figure}
Figure~\ref{fig:datasetsDistribution} shows the distribution of shape errors $E_{aver}$ and $E_{max}$ on all samples in the test dataset. It is obvious that our method is superior to the other two methods. Specifically, when checking the CDF curve of $E_{aver}$, our network can lead to a result of 94\% test samples having an error less than $1.0$~cm. Only 63\% test samples can result in a reconstruction with $E_{aver}$ less than $1.0$~cm by using the approach of Ji et al.~\cite{ji2019human}, and all human models reconstructed by the method of Dibra et al.~\cite{dibra2016hs} have the error $E_{aver}$ larger than $1.0$~cm. The maximal $E_{aver}$ of our approach on test samples is $3.4$~cm, which is also much smaller than the other two methods.
In terms of the $E_{max}$'s CDF curves, 84\% of test samples can reach the maximal error less than $4.0$~cm by using our method. However, the other two methods can only have 70\% and 18\% reach this level of accuracy. The histograms of frequency distribution for the $E_{aver}$ and $E_{max}$ errors shown in the left of Fig.\ref{fig:datasetsDistribution} also lead to the conclusion that most of our reconstructed models fall in the region with smaller errors. The average errors are also listed and compared in Table \ref{tab:avegErrComparison}. Besides of the statistical results, we also show the accuracy of reconstruction on individual models that have the largest average error $E_{aver}$ and the largest maximal error $E_{max}$ respectively. Three worst cases are displayed in Fig.\ref{fig:examples} by using the colorful error map. The results from \cite{dibra2016hs,ji2019human} and ours are listed by using the same range of errors (Unit: centimeter). \revise{}{The color distribution changing from blue to red is employed to denote the variation of geometric error from zero to maximum.} Again, our approach performs the best.

\begin{table}
\caption{Quantitative comparisons on three semantic curves with statistical errors evaluated on all $862$ models in the test dataset. The mean and the standard deviation of absolute errors (Unit: centimeter) are denoted as $\mu$ and $\sigma$. The best results are highlighted by bold fonts.}
\label{tab:real3DError} \vspace{5pt}
    \centering \small
    \begin{tabular}{c|c|c|c}
    \hline
$(\mu,\sigma)$ & Dibra et al.~\cite{dibra2016hs} & Ji et al.~\cite{ji2019human} & Our Approach \\
    \hline
    \hline
    {\textbf{Hip} } & (4.09, 3.86)    & (1.44, 1.34)   & \textbf{(0.67, 0.66)} \\
    \hline
    {\textbf{Waist} }  & (6.26, 4.21)     & (1.66, 1.70)   & \textbf{(0.76, 0.81)} \\
    \hline
    {\textbf{Chest} }  & (4.99, 4.03)     & (1.47, 1.40)   & \textbf{(0.81, 0.69)} \\
    \hline
    \end{tabular}
    \label{tab:girthErrorDist}
\end{table}

In addition, we also evaluate the girth errors (B:Chest, W:Waist and H:Hip) on all models in our test dataset. Our results are compared to the other two approaches \cite{dibra2016hs,ji2019human}. The distributions of girth errors are given in Fig.~\ref{fig:GirthErrorDistribution}. The error statistics are reported in Table~\ref{tab:girthErrorDist}. It can be observed that our method performs the best. The errors of our results on these three girths are always less than one centimeter.


\section{Application}\label{secApplication}
In this section, we discuss practical issues to use our network of human shape modeling in an Industry 4.0 application -- design automation for customized clothes. First, a learning transfer approach is applied to solve the problems of dataset with small number of 3D human models. After that, we demonstrate a smartphone APP that is based on our network to reconstruct 3D human models for generating customized clothes.

\subsection{Transfer learning}
\begin{figure}[t]
\centering
{\includegraphics[width = \linewidth]{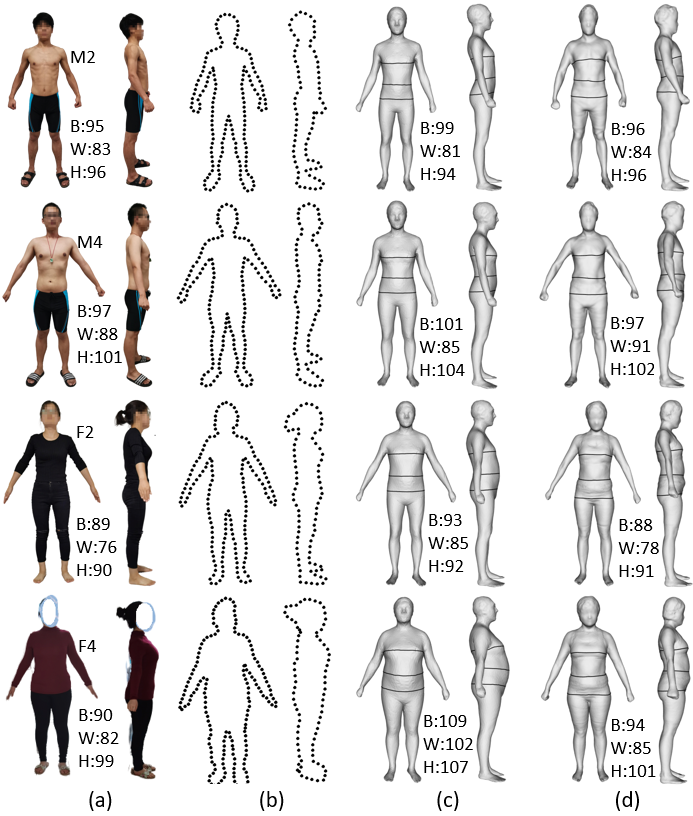}}
\caption{Reconstruction results by the network before vs. after transfer learning -- examples are obtained from the real photos of user study: (a) the original images with background removed (ground truth of chest, waist and hip girths are given), (b) the sampled silhouette points, (c) results obtained by using the original network (i.e., before transfer learning), and (d) results obtained by the network after transfer learning. From these results, we can find that the network updated by transfer learning can capture more female / male features -- i.e.,  gender-oriented patterns can be captured more clearly.}\label{fig:transfer1}
\end{figure}

By using a large dataset of human models released in~\cite{pishchulin2017building}, we are able to train a regression model for estimating the 3D shape of human model in high accuracy. However, this dataset is not able to reflect the gender-oriented (or geographical) shape patterns. Ideally, different datasets will be needed for training networks for different genders or races. As shown in Fig.~\ref{fig:transfer1}(c), the human model reconstructed by the network trained on a database with mixed genders (i.e.,~\cite{pishchulin2017building}) does not show a strong pattern of gender.

In order to avoid collecting large dataset of 3D human models, we apply the strategy of transfer learning~\cite{torrey2010transfer} to obtain gender-oriented networks from datasets with limited number of samples. Two \revise{}{registered} small datasets with 31 female models and 40 male models, \revise{}{released by Hasler et al.~\cite{hasler2009statistical},} respectively are conducted for the transfer learning. Specifically, we fix the parameters of all blocks of BlockA and BlockB in our network, and only update parameters in the fully connected layers (see the yellow FC layers in Fig.\ref{fig:architecture}) by using these small datasets. Again, each training sample is still represented by two silhouette images and the corresponding shape coefficients $\varphi_s$. We still conduct the similar strategy introduced in Section~\ref{secTech} to generate our new training and test datasets by using $1.2 r_{min}$ as threshold, $K=11$ for neighborhood search and uniformly inserting three samples between two neighboring samples, which results in two training databases with 425 models for female and 472 models for male. \revise{}{Table \ref{tab:compTwo} shows the errors of feature curves between the predicted results and their real ones. We can find that most mean absolute errors are less than three centimeters.} As a byproduct of transfer learning, the human models employed can have mesh connectivity different from the samples used for training the original network.

\begin{table}
\caption{\revise{}{Error estimation between the predicted feature curve and the real one in the corresponding test datasets. The mean absolute error and the standard deviation ($\mu$, $\sigma$) are given (Unit: centimeter).}}
\label{tab:errorOnSmallSets} \vspace{5pt}
    \centering \resizebox{\linewidth}{!}{
    \begin{tabular}{c|c|c|c}
    \hline
    Error ($\mu$, $\sigma$)    & Hip      &  Waist &  Chest \\
    \hline
    \hline
    Female & (2.02, 1.58)     & (1.48, 1.21)  & (2.01, 1.60) \\
    \hline
    Male  & (2.52, 2.11)    & (3.83, 2.38)   & (2.48, 2.32) \\
    \hline
    \end{tabular}}
    \label{tab:compTwo}
\end{table}

\begin{table}
\caption{Comparison of results generated by the networks before vs. after transfer learning by using the real photos of user study \cite{liu2020userstudy}. The mean absolute error and the standard deviation ($\mu$, $\sigma$) are given (Unit: centimeter).}
\label{tab:real3DError} \vspace{5pt}
    \centering \resizebox{\linewidth}{!}{
    \begin{tabular}{c|c|c|c}
    \hline
    Error ($\mu$, $\sigma$)    & Hip      &  Waist &  Chest \\
    \hline
    \hline
    Original network & (3.16, 2.45)     & (5.83, 4.93)  & (4.13, 2.41) \\
    \hline
    With transfer learning  & \textbf{(2.76, 2.74)}    & \textbf{(3.72, 2.84)}   & \textbf{(2.30, 2.47)} \\
    \hline
    \end{tabular}}
    \label{tab:comparisonWithTransferLearning}
\end{table}

We apply the real photos obtained in user study (\revise{Sec.}{i.e., Fig.1 of \cite{liu2020userstudy}}) to the new network obtained from transfer learning. The measurement errors in three girths are evaluated and compared with the results generated by the network without transfer learning (see Table~\ref{tab:comparisonWithTransferLearning}). Besides of these statistical analysis, examples of reconstruction are shown in Fig.~\ref{fig:transfer1}. As can be found from Fig.~\ref{fig:transfer1}(d), the gender-oriented patterns and measurement accuracy have been significantly enhanced after apply the transfer learning to our network with the help of a gender-separated small training dataset.

\begin{figure}[t]
\centering
\includegraphics[width = \linewidth]{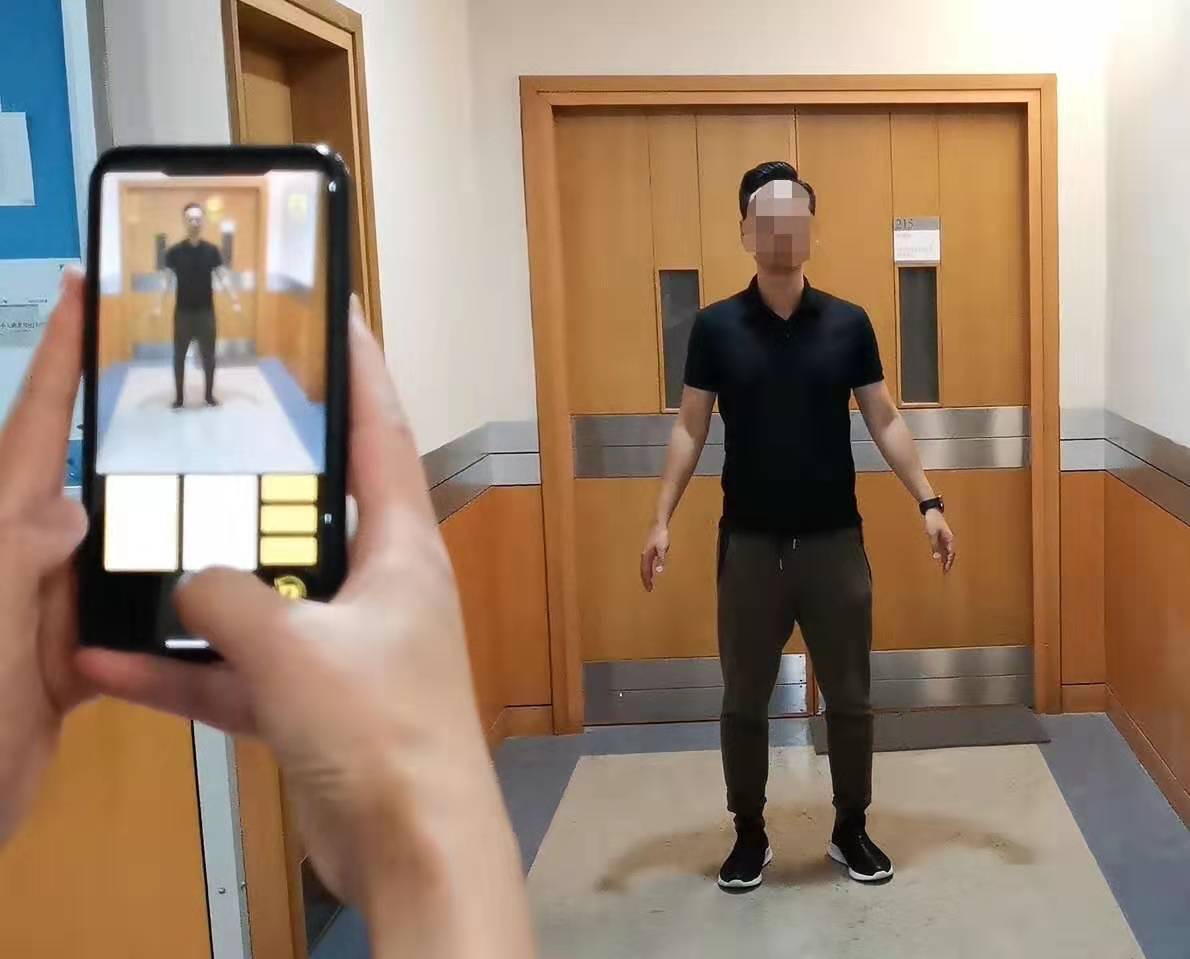}
\caption{The scenario of using our smartphone APP for 3D human model reconstruction.}\label{fig:phoneAPPScenario}
\end{figure}
\begin{figure}
\centering
\includegraphics[width = 4cm]{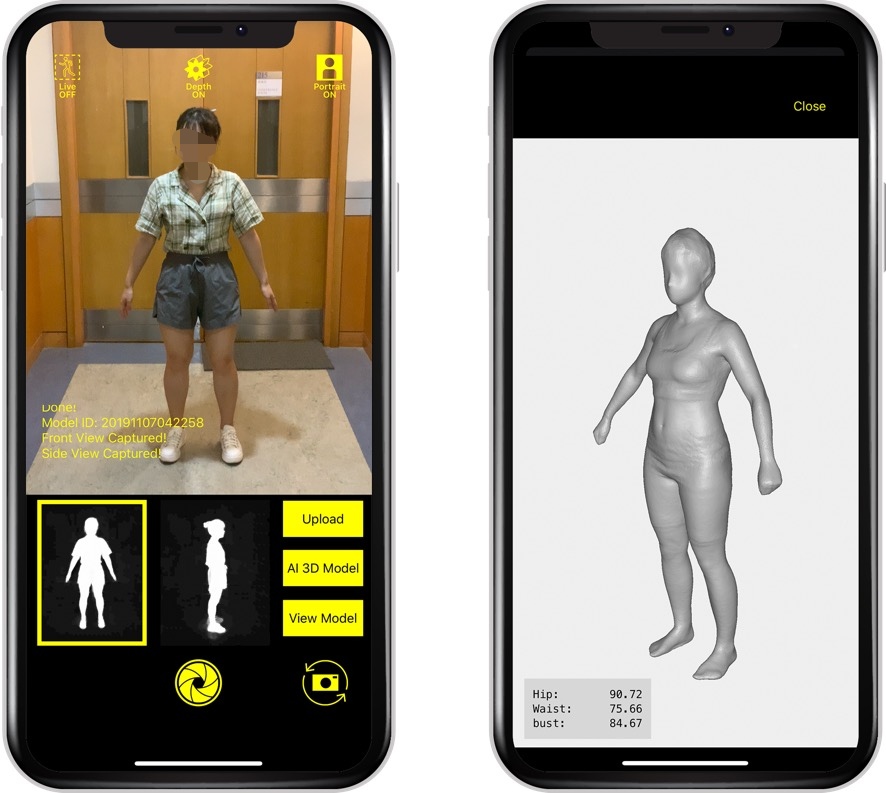} \hspace{8pt} \includegraphics[width = 4cm]{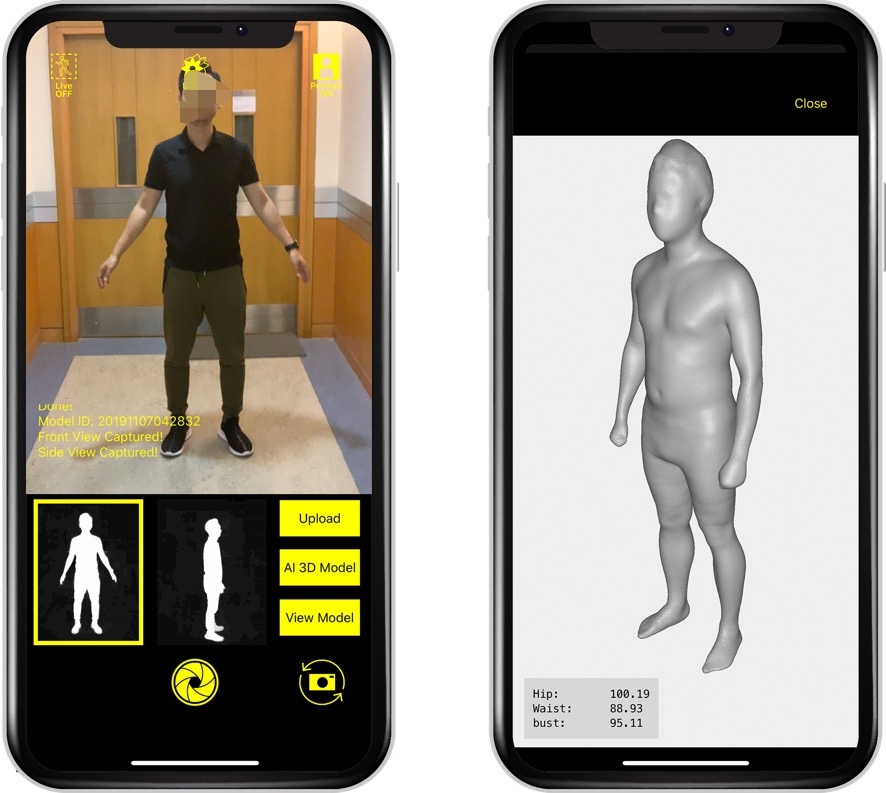}
\caption{Interface of our smartphone APP.}\label{fig:phoneDisp}
\end{figure}

\subsection{Customized design}
\begin{figure}\centering
\includegraphics[width = \linewidth]{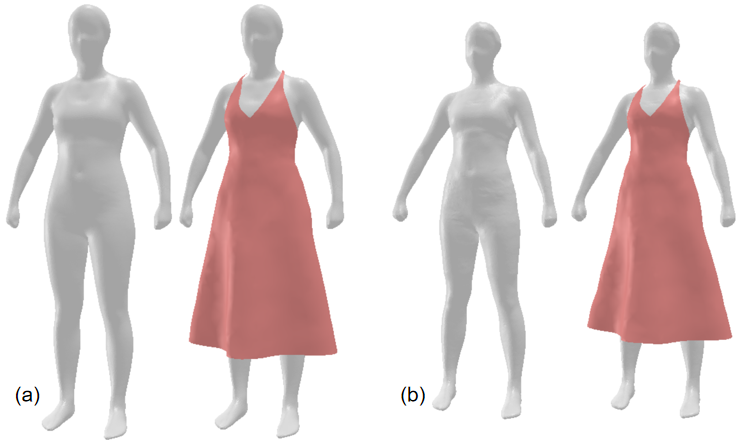}
\caption{\revise{}{The dress originally designed for a standard model (a) can be automatically transferred to a customized shape perfectly fit the 3D human body generated from our network (b).}}\label{fig:designTransfer}
\end{figure}

Based on the network proposed in this paper, we have developed a smartphone APP to reconstruct 3D human models from the photos of individuals -- see the scenario demonstrated in Fig.\ref{fig:phoneAPPScenario} and the APP interface shown in Fig.\ref{fig:phoneDisp}. Background is removed by using the function that is built in the smartphone. Note that, a strategy similar to \cite{Wang2003CADFuzzy} is employed to supervise the process of taking photos for the side view of a human model (i.e., by checking the completeness of legs overlap). A video demonstration of this APP can be found at:~\url{https://youtu.be/JEPAmiB0wYI}. By using the barycentric coordinate, the feature curves predefined on a human model with standard shape can be automatically generated on the reconstructed models for the measurement purpose. Moreover, the perfect fit clothes for individual customers can be automatically generated by using the design transfer technology\footnote{\revise{}{An implementation of this design transfer on 3D human models can be access at:~\url{https://zishun.github.io/projects/3DHBGen/}.}} \cite{wang2007volPara,meng2012designTrans}. \revise{}{Figure \ref{fig:designTransfer} has demonstrated the design transform result by the female models reconstructed from two silhouette images.}
%

\section{Conclusions and Future Work}
In this paper, we have developed a novel architecture of CNN for modeling 3D human bodies from two silhouette images. Our network is concise and efficient, which is benefit by developing an architecture with fusion blocks and also conducting samples on silhouette as input. More accurate models can be generated by our network with only 2.4M coefficients. The learning of our network is first conducted on samples obtained by augmenting a publicly accessible dataset, and the transfer learning method is applied to make it usable when a dataset with only small number of human models is available. Experimental tests have been conducted to prove the effectiveness of our network.

The major limitation of our current work is that the input sample points should capture the shape of human silhouette accurately. This requirement is too strong in some scenarios of practical usage -- e.g., the long hair of female users may lead to incorrect shape at the region of neck. \revise{}{In addition, we argue that integrating 3D information into our network or modifying the structures of BlockA and BlockB, may further improve the accuracy of shape prediction which will be explored in our future work.} Lastly, the input side-view should have the orientation consistent to the training samples. We plan to eliminate this requirement in our future development (i.e., make network automatically adaptive to the left or the right-views as input). Moreover, as our method performs the best in neutral poses, we plan to enhance the smartphone APP by providing the function of pose similarity guidance based on skeleton extraction.

\section*{Acknowledgement}
Part of this work is completed when B. Liu and C.C.L. Wang worked at the Chinese University of Hong Kong. This work is partially supported by HKSAR Innovation and Technology Commission (ITC) Innovation and Technology Fund (Project Ref. No.: ITT/032/18GP), the Natural Science Foundation of China(61976040, 61702079, 61762064), and the Science and Technology Development Fund of Macau SAR (File No.: SKL-IOTSC-2018-2020, 0018/2019/AKP and 0008/2019/AGJ). The authors also would like to thank the valuable comment given by Zishun Liu in private communications.

\bibliography{referencess}

\end{document}